\documentclass[sn-apa, iicol]{sn-jnl}% APA Reference Style
\jyear{2022}%

\theoremstyle{thmstyleone}%
\theoremstyle{thmstyletwo}%

\theoremstyle{thmstylethree}%

\begin{document}

\title[Article Title]{Modelling customer churn for the retail industry in a deep learning based sequential framework}


\author[1]{\fnm{Juan Pablo} \sur{Equihua}}\email{je18890@essex.ac.uk}

\author[1,3]{\fnm{Henrik} \sur{Nordmark}}\email{henrikn@profusion.com}

\author[2]{\fnm{Maged} \sur{Ali}}\email{maaali@essex.ac.uk}

\author*[1,4]{\fnm{Berthold} \sur{Lausen}}\email{berthold.lausen@fau.de}


\affil[1]{\orgdiv{Department of Mathematical Sciences}, \orgname{University of Essex}, \orgaddress{ \city{Colchester}, \country{United Kingdom}}}

\affil[2]{\orgdiv{Essex Business School}, \orgname{University of Essex}, \orgaddress{ \city{Colchester}, \country{United Kingdom}}}

\affil[3]{\orgdiv{Innovation}, \orgname{Profusion Media Ltd.}, \orgaddress{\street{Paul Street}, \city{London}, \country{United Kingdom}}}

\affil[4]{\orgdiv{Institute of Medical Informatics, Biometry and Epidemiology, School of Medicine}, \orgname{Friedrich-Alexander University Erlangen-Nuremberg},
    \orgaddress{\street{Waldstr. 6}, \city{Erlangen}, \postcode{91054},  \country{Germany}}}

%\affil[2]{\orgdiv{Essex Business School}, \orgname{University of Essex}, \orgaddress{\street{Wivenhoe Park}, \city{Colchester}, \postcode{CO4 3SQ}, \state{Essex}, \country{United Kingdom}}}


\abstract{As retailers around the world increase efforts in developing targeted marketing campaigns for different audiences, predicting accurately which customers are most likely to churn ahead of time is crucial for marketing teams in order to increase business profits. This work presents a deep survival framework to predict which customers are at risk of stopping to purchase with retail companies in non-contractual settings. By leveraging the survival model parameters to be learnt by recurrent neural networks, we are able to obtain individual level survival models for purchasing behaviour based only on individual customer behaviour and avoid time-consuming feature engineering processes usually done when training machine learning models. }

%%================================%%
%% Sample for structured abstract %%
%%================================%%

% \abstract{\textbf{Purpose:} The abstract serves both as a general introduction to the topic and as a brief, non-technical summary of the main results and their implications. The abstract must not include subheadings (unless expressly permitted in the journal's Instructions to Authors), equations or citations. As a guide the abstract should not exceed 200 words. Most journals do not set a hard limit however authors are advised to check the author instructions for the journal they are submitting to.




\maketitle

\section*{Introduction}\label{sec1}

Finding innovative methods to mitigate customer churn and improve retention has historically been an active task for businesses across a wide range of sectors such as finance, technology, banking, insurance, among others. Particularly, in the retail sector, different studies such as the ones conducted by  \cite{Reichheld_1990,  VANDENPOEL2004196} have shown that acquiring new customers is usually between 5 to 12 times more expensive for companies than retaining existing ones. Although this ratio varies across companies. Marketers recognise several advantages of dedicating major efforts to identify which customers are likely to churn in order to design more competitive marketing strategies for customers. \cite{Reichheld_1990} showed that an improvement of just 5\% in customer retention leads to an increase of 85\% in profits for the banking sector, 50\% for insurance brokerage, and 30\% in the automotive industry. Unfortunately, the underlying reasons that lead customers to churn might vary for different businesses and industries, although marketers have identified that bad customer service, poor value proposition, low-quality communications, and lack of brand engagement as the main four reasons for voluntary customer churn.

A common approach used by companies to identify future churning customers in non-contractual settings such as retail, where customers are not subject to a subscription model and can change their purchasing habits without informing the company, is by using statistical and machine learning methods to predict which customers are likely to stop doing business with the company within a certain time window. Then, individuals with high probability of churning can be targeted in one or several retention campaigns which commonly offer product promotions specifically designed to provide an incentive to customers to make a purchase and keep them engaged with the brand, as noted by \cite{Borah_2019}. While the concept of predicting customer churn is relatively intuitive, the realities involved in designing such systems can be challenging for multiple reasons. Firstly, in a non-contractual setting such as retail, customers can change their purchasing habits at any moment, and typically the longer a customer takes to make their next purchase, the lower the probability is of that customer returning at all. Secondly, the use of multivariate statistical methods and machine learning techniques involves the extraction of hand-crafted characteristics for each customer that are usually proposed by subject matter experts. The quality, quantity, and type of these characteristics will have a direct impact on the final performance of any classification method used to find churning customers \cite{bengio2014}. Furthermore, as customers' behaviour differ in different sectors and in different companies, finding a relatively good set of characteristics that generalizes well across companies may be difficult to find and would thus lead one to the time-intensive task of finding such characteristics in each new context.

To overcome these issues, several research have explored the use of methods such as artificial intelligence,  artificial neural networks, and representation learning to obtain customer representations without the need of human intervention, with the main goal of avoiding the time-consuming feature engineering step that companies need to carry while designing machine learning models and assuming that AI will change marketing strategies and customer behaviours over the next decade as noted by \cite{Davenport_2019}. For instance, \cite{Spanoudes2017} use abstract feature vectors in a 4-layer neural network to predict which customers are likely to churn in a monthly defined horizon. However, this and similar methods like the ones proposed by  \cite{Coussement2008, Hung2006, Tamaddoni2010} do not take into account that the event of interest might have not happened yet for all individuals at observations period, as it is considered in survival-based techniques proposed in the literature, such as the Kaplan-Meier estimate that has also been used to predict customer churn due to their inherent design to model probabilities of an event occurring over a lapse of time without the extra complexity of obtaining large amounts of features to represent customers as in \cite{JAMAL2006, Wong2011, GUL_2020}. 

To effectively predict customer churn and design targeted marketing campaigns, it is important to design more effective methods that consider both, the inherent purchasing behaviour of individual customers and the censoring effect induced by the uncertainty of not being capable to identify customers that have already ended their relationship with the business against the ones that simply are during a pause between transactions. Traditional techniques to predicting which customers are likely to churn soon, usually approach just one of these two desired characteristics. 

This research suggests a new approach for modelling customer churn in non-contractual settings, such as the retail industry, with three main goals. Firstly, outperform traditional machine learning and survival-based methods that use hand-crafted features extracted from behavioural information of customers, this is achieved by replacing the feature engineering phase with a deep learning-based approach that performs model parameters estimation with the use of recurrent neural networks. Secondly, reduce the inherent algorithmic bias in marketing models by excluding customer information such as demographic data and define customers’ behaviour entirely from the time customers' purchases take to occur. Finally, obtain reliable time-to-event models capable of capturing the real buying patterns of customers over time by obtaining individual-level distributions of each customer's arrival times.

The rest of the paper is organized as follows, second section reviews the current literature of customer churn prediction with deep learning methods and survival analysis. Third section outlines the current methodology and data pre-processing required for the analysis. Fourth section presents the experimental results in a real transactional dataset provided by a large retailer in the UK. Finally, conclusion and future work is provided in the fifth section .

\section*{Literature Review}\label{section2}
\subsection*{Survival Analysis and Time-to-event modelling}\label{section2_1}
Survival analysis (SA) is the field of statistics focused in modelling time-to-event data over future lifespans,  i.e.   estimating the probability  of  an  event  occurring  beyond  a certain  time  in  the  future.   In  contrast  to  Machine Learning (ML)  methods  such  as  regression and  tree-based  models,  where all events have already occurred at observation time, survival analysis assumes that the event might not have happened at the time of evaluation for some individuals, but it could happen in the future if the observation period were to be extended, this effect is known as right censoring. Survival analysis estimates the probability of the outcome event not occurring up to a time $t$ and accounts for the presence of censoring in data with a survival function $S(t)$ defined as 

\[
S(t)=P(T>t)=1-F(t),
\]

where $T$ is a random variable defined from the distribution of events over time, and $F(t)$ is the cumulative distribution of the event times, which is usually  modelled with respect to a set of subject attributes $X$ (a.k.a covariates, or predictors, or features), i.e., $S(t\mid X)=P(T>t \mid X)=1-F(t \mid X)$.  For a survival function $S(t)$, the hazard ratio, or hazard function $\gamma(t)$ defines the event rate at time $t$, if the event has not occurred up to time t the hazard functions is expressed as 

%\[
%h(t) = \lim_{\vartriangle t \to 0} \frac{1}{\vartriangle t}P(T < t +\vartriangle t | T %\geqslant t ) = \frac{S^\prime (t)}{S(t)}
%\]

\[
\gamma(t) = \lim_{\vartriangle t \to 0} \frac{P(T < t +\vartriangle t \mid T \geqslant t )}{\vartriangle t} = \frac{S^\prime (t)}{S(t)},
\]

with $S^\prime (t) = -f(t)$ and $f(t)$ probability density function of the events time distribution.

Survival analysis methods can be classified into three main different categories: 1) Parametric methods, which assume a specified distribution of survival times as well as a functional form for model covariates. 2) Semi-parametric methods, which enforce a functional relation between covariates and the survival function, but do not impose a specific form of the hazard function. And 3) Non-Parametric methods, which do not impose any assumption on the survival function nor the covariates distributions. Out of all different techniques proposed in the literature, two common methods used in industry applications are the Cox Proportional Hazard (CPH) model \cite{cox1972} and the \cite{kaplan_meier} estimator due to their flexibility and easiness process at implementation. The CPH semi-parametric form allows to linearly combine distributions from multiple covariates with a baseline hazard to obtain time-dependant probability estimates of individuals' risk at time t. Whereas the Kaplan-Meier non-parametric structure allows companies and researchers to have a robust baseline and reliable predictions of individuals’ risk over time in an easy and scalable way. The Kaplan-Meier estimator $\hat{S}(t)$ of the survival function of individuals is defined by 

\[
\hat{S}(t) = \prod_{k:t_{k}<t} \left(1-\frac{d_{k}}{n_{k}} \right), 
\]
where $d_{k}$ is the number of individuals that experienced the event at the time $t_{k}$ and $n_{k}$ is the total number of individuals at risk at time $t_{k}$.

Survival models have been widely used by several companies and marketers around the world to predict customer churn due to their simplicity and flexibility to include multiple covariates into the hazard function estimation. \cite{VANDENPOEL2004196} explored the use of proportional hazards to model customer attrition in European financial services. \cite{Wong2011} used the Cox regression to identify demographic and temporal covariates that impact customer retention in a telecommunication company with base in Canada. \cite{JAMAL2006} linked time-dependant covariates from customer service, payments, and recovery systems with the use of a Weibull hazard to identify churning customers for a satellite TV service in South America. \cite{Mavri2008} examine potential predictors in customer switching behaviour for the Greek banking sector. 

However, CPH does not directly model survival probabilities, but the hazard function of individuals at time $t$ as $\gamma(t\mid X)=\gamma_{0}(t)exp(X^T\beta)$, which is the probability that an individual will experience the event of interest within a time interval given that the event has not happened up to the beginning of the interval, and it is obtained from a baseline hazard that only depends on time $\gamma_{0}$, and a time-independent function obtained from the individual's covariates as $X^T\beta$, where $\beta$ is the $n\textnormal{-dimensional}$ weights vector associated to each individual covariate. Once the hazard function is known, the survival function can be retrieved with the use of the cumulative hazard function $\Gamma(t)=\int_{0}^{t}\gamma(s)ds$, as $\hat{S}(t)=\text{exp}(-\Gamma(t))$. 

Typically, the CPH model is fitted in two steps \cite{kvamme2019}. Firstly, the parametric part of the model that only depends on individual's covariates $exp(X^T\beta)$ is fitted by maximising the Cox partial likelihood, as it does not depend on the baseline hazard function $\gamma_{0}$, then, the non-parametric baseline is estimated based on the parametric results obtained in the previous step.  

Although survival techniques have proofed their efficiency to predict customer churn accurately in several applications, these methods also have their drawbacks. Firstly, these techniques commonly assume that the event of interest can occur only once and it will happen with probability of one if individuals are observed for enough time, which is not the case in modelling customer behaviour where inherently individuals can make several purchases over their lifetimes \cite{Mavri2008, Spanoudes2017, Tamaddoni2010} or not come at all again after certain time. A pragmatic approach to model these issues is by resetting the customers' survival probability to 1 immediately after they make a new purchase and introduce a cure factor which represents the probability that individuals will not make any further purchase \cite{Amico20218}.

Secondly, performance assessment of survival models applied to purchasing behaviour might not be straightforward, as survival techniques estimate  the probability of events occurring over time, whereas evaluating traditional customer purchases is modelling a binary classification problem ‘customer will make a purchase eventually vs customer will not make a purchase ever again’. Allowing solutions which can be optimal point-wise, but not overall.
% Leading into a potential evaluation bias at deciding which model is best to predict customer churn. 

Finally, CPH still enforce the use of a constant hazard function for all individuals, which is an unrealistic assumption at modelling purchases for the non-contractual retail industry, where customers can change their buying patterns at any time. Different studies have looked to overcome this challenge by using mixture models on top the original Cox model \cite{nagpal_2019}, by training CPH models in adversarial frameworks \cite{chapfuwa2018} or using time-dependant hazard functions \cite{Fisher_1999},  or combining survival models with network-based architectures, for instance, \cite{Nagpal2020} propose a fully parametric mechanism called \emph{Deep Survival Machines} to learn non-linear representations of covariates without the need of the strong hazard assumption. Whereas \cite{Ren2018} propose a deep recurrent survival model to predict the likelihood of an event without assuming any specific distribution on the survival function while accounting censorship presence in data. 

\subsection*{Neural Networks in Survival Analysis}
\subsubsection*{Artificial Neural Networks}
Artificial Neural Networks (ANN) are popular machine learning models capable of learning complex non-linear patterns present in data with the use of nodes and weighted connections interrelated in an architecture design that is mainly inspired in the structure of the human brain. Neural networks have proof to be models capable of outperforming several machine learning techniques such as logistic regression and tree-based models \cite{Menghani2021} in different domains, such as Natural Language Processing (NLP) \cite{Camacho2017}, computer vision \cite{CNNs}, and machine translation \cite{sutskever_2014}. ANN have been also widely explored in the fields of survival analysis by \cite{kvamme2019} and churn prediction by \cite{Sharma2013}, for tasks where the observations may allow to presence of censoring and covariates can be extracted from the input data.

Mathematically, a neural network ($NN:\mathbb{R}^n \rightarrow \mathbb{R}^m$) with $n\textnormal{-dimensional}$ input and $m\textnormal{-dimensional}$ output can be seen as a linear combination or a function of $X \in \mathbb{R}^n$ features (nodes) with their corresponding weights $W \in \mathbb{R}^{n \times m}$ and bias term $b \in \mathbb{R}$ (connections), followed by a non-linear activation function like the sigmoid or hyperbolic tangent functions, i.e., $NN(X) = \varphi(b + W^{T} X)$, where  $\varphi:\mathbb{R}^m \rightarrow \mathbb{R}^m $ represents the chosen activation for the network. In practice, deep neural network architectures contain multiple nodes in their input and intermediate layers to allow the learning of complex non-linear mappings from input to output data during the training process, which is performed by minimising a loss function of the network outputs and the targets with the use of the back-propagation algorithm proposed by \cite{Rumelhart1987} or one of its variants. In order to simplify the notation, we denote $NN_{X}$, as the neural network learnt from training data $X$. 

% as the neural network with input $X$ and weights and bias vector $\theta$, for all the model's layers, where $\theta = [\theta_{1},\theta_{1},...,\theta_{j}]$, and $\theta_{j} = (W_{j}, b_{j})$ represents the weights and bias for layer $j$, each layer $l_{j}$ in the network can be written as $l_{j} = \varphi_j(b_{j}+ W^{T}_{j} X_{j})$.

% $l_{j} = \sum_i(w_{j,i}x_{i}, +w__{j,0})$.

Although neural networks are powerful machine learning models, these also have their caveats. Firstly, due to their structure with usually multiple intermediate layers, the model might contain thousands or millions of parameters that need to be optimized in the training process, so these techniques require large volumes of input data at training to avoid over-fitting and provide reliable predictions. Secondly, neural networks typically provide only point estimates of the target variable and do not capture the uncertainty in their predictions, which is not favourable when the goal is estimating a probability distribution such as in this work. A common way to approach these issues is by taking a probabilistic approach and assume a distribution over the target variable $Y$, i.e., $Y \sim Q(\theta)$ with unknown parameter vector $\theta$, then the neural network outputs are only estimates of the parameter distribution instead of estimates of the target variable, i.e, $NN_{X}(x) = \varphi(b + W^{T} x) = \hat{\theta}$, and target point estimates can be obtain as the expectation of the distribution w.r.t. the input data as $\hat{y} = E[Q(\hat{\theta} \mid x)]$. Naturally, this process can be implemented not only for the output layer, but in each or some of the intermediate layers as well, leading into the field of Bayesian neural networks that are trained by minimizing the Kullback-Leibler divergence between the estimated posterior distribution $Q(\hat{\theta})$ and a defined prior distribution as noted by \cite{LAMPINEN2001}. 

\subsubsection*{Recurrent Neural Networks}
\begin{figure}[!htbp]
\centering{}
\includegraphics[scale=0.6]{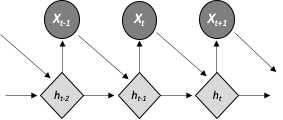}
\caption{Graphical representation of the hidden state structure of Recurrent Neural Networks.}
\label{fig:RNN_structure}
\end{figure}

Recurrent neural networks (RNN) are an extension of traditional network architectures widely used in sequential modelling applications such as text classification by \cite{yi2003}, language modelling, and machine translation by \cite{sutskever_2014}. Due to their internal structure, RNN can encode information of time-dependant random variables $X_{t}$ into a hidden state $h_{t}$, which is then used to define the probability distribution of the time-dependant target variable $Y_{t}$ as  $P(Y_{t} \mid h_{t})$. In RNN, the hidden state $h_{t}$ evolves over time by using information of $X_{t}$ at each time-step and combining it with the previous hidden state with a function $g$, i.e., at each time-step $h_{t}=g(h_{t-1},X_{t})$ as shown in Figure \ref{fig:RNN_structure}. 

However, the choice of the encoding function $g$ is not straightforward, as it needs to be capable of capturing long-term dependencies in sequential data and not over-fit during the training process. Popular choices for $g$ are functions so called 'memory cells' such as Long-Short Term Memory (LSTM) introduced by \cite{LSTMs} and Gated Recurrent Units presented by (GRU) \cite{GRUs}, which use an internal gated mechanism based with sigmoid functions to store and forget information through the recurrent training process. \cite{bennis2020} highlight the benefits of using recurrent neural networks (RNN) over any other type of models for predicting customer churn where covariates extracted from data vary over time.

Recent research have proposed methods to combine RNN with survival analysis to outperform traditional CPH models and its variants. In these methods, typically the survival model is fully parameterised with the output of a recurrent network to predict the empirical distribution of future events and make use of time-dependant covariates. \cite{Giunchiglia2018} proposed RNN-SURV for the medical field, this method takes characteristics of patients over a period of time, at each time-step the model computes both, the risk score, and the survival function of each patient in a personalised manner. \cite{Martinsson2016} proposed WTTE-RNN (Weibull Time-to-Event RNN) which consist mainly in the use of a recurrent network to estimate the parameters of a Weibull distribution, then this estimated Weibull distribution is used to predict engines time-to-failure in the field of machinery maintenance. \cite{Chen2018} proposed MAT-RNN (Multivariate Arrival Times RNN) to extend prediction of survival frameworks to multiple arrivals setting, such as prediction purchases in demand forecasting. \cite{bennis2020} proposes a recurrent architecture to model the parameters in a mixture Weibull distribution for time-to-event analysis. 

\subsection*{Asymmetric loss functions}

Typically, the training process of neural networks is based in adjusting the networks’ weights and biases via the back-propagation to find the weights that minimise the error between model’s predictions $\hat{y}$ and real observed values $y$ with respect to a loss function $L(\hat{y}, y): \mathbb{R}^m \times \mathbb{R}^m \rightarrow \mathbb{R}$ , which measures the discrepancy between predictions and target values. In classification or regression tasks the target variable is fully known and given by a set of true labels $y$, thus the model predicts $\hat{y} = NN_{X}(x)$ and the model's error can be obtained straightforwardly. Popular choices of loss function are the Mean Squared Error ($MSE = \frac{1}{N}\sum_{i=1}^{N}(y_{i}-\hat{y}_i)^2$), and Mean Absolute error ($MAE = \frac{1}{N}\sum_{i=1}^{N} \mid y_{i}-\hat{y}_i \mid$) for regression, and the binary or categorical cross entropy ($Entropy=-\sum_{i=1}^{N}y_{i} \cdot log (\hat{y}_{i})$) for classification. However, these loss functions do not consider the presence of censored labels in the data and only provide reliable estimations when over-predicting or under-predicting the real value of the target does not have a significant impact during model training, which is not the case at estimating the parameters of a time-to-event distribution.

The estimation of parameters for an exponentially shaped time-to-event distributions has been widely explored over the last decades, as several life problems such as waiting time problems or time intervals between events usually distribute similar to an exponential shape.  Several authors, such as  \cite{Zellner_1986, Varian_1975, Srivastava_2007} have shown that the use of asymmetric loss functions outperform the estimation of parameters carried with quadratic-type losses such as MSE and MAE, due to their inherent structure that consider both, the goodness of fit against the distribution to estimate and the precision of individual estimations. In addition, generalized linear models \cite{Nelder_1972} provide a statistical framework to model 
non-symmetric distributions.    

Popular choices heuristically motivated for asymmetric loss functions are the LINEX loss function \cite{Varian_1975}, which increases exponentially on one side of zero and linearly in the other side. Similarly, balanced loss functions (BLF) \cite{Zellner_1986} combine the distance of a given estimator to the target distribution and to its unknown parameters. Balanced loss functions are usually considered in the field of Bayesian statistics as these may consider prior knowledge of parameters that can be captured in the form of a prior distribution.  
% [EXPAND ANALYSIS FOR ASYMMETRIC LOSS FUNCTIONS. THERE IS ONLY 3 REFERENCES QUOTED.]

\section*{Methodology} \label{section3}
This section describes the mathematical representation of customer transactions data and the model architecture used to estimate how likely individuals are to make purchases over time. We aim to train a neural network capable of estimating the parameter of exponentially distributed arrival times by using the information of previous observed and censored event times. Then, this neural network can be used to predict the next event time and therefore, estimate the survival distribution at customer level. 

\subsection*{Data Representation} \label{data_representation}
\begin{figure}[!htbp]
\centering{}
\includegraphics[scale=0.45]{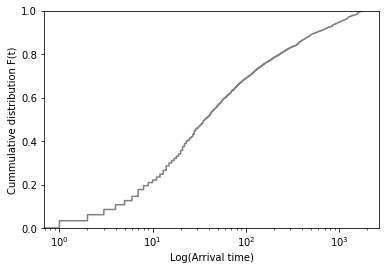}
\caption{Distribution of time between consecutive customer purchases $t_{k , \cdot}$ in logarithmic scale.}
\label{fig:itt_distribution}
\end{figure}

\begin{figure}[!htbp]
\centering{}
%\captionsetup{width=.8\linewidth}
\includegraphics[scale=0.35]{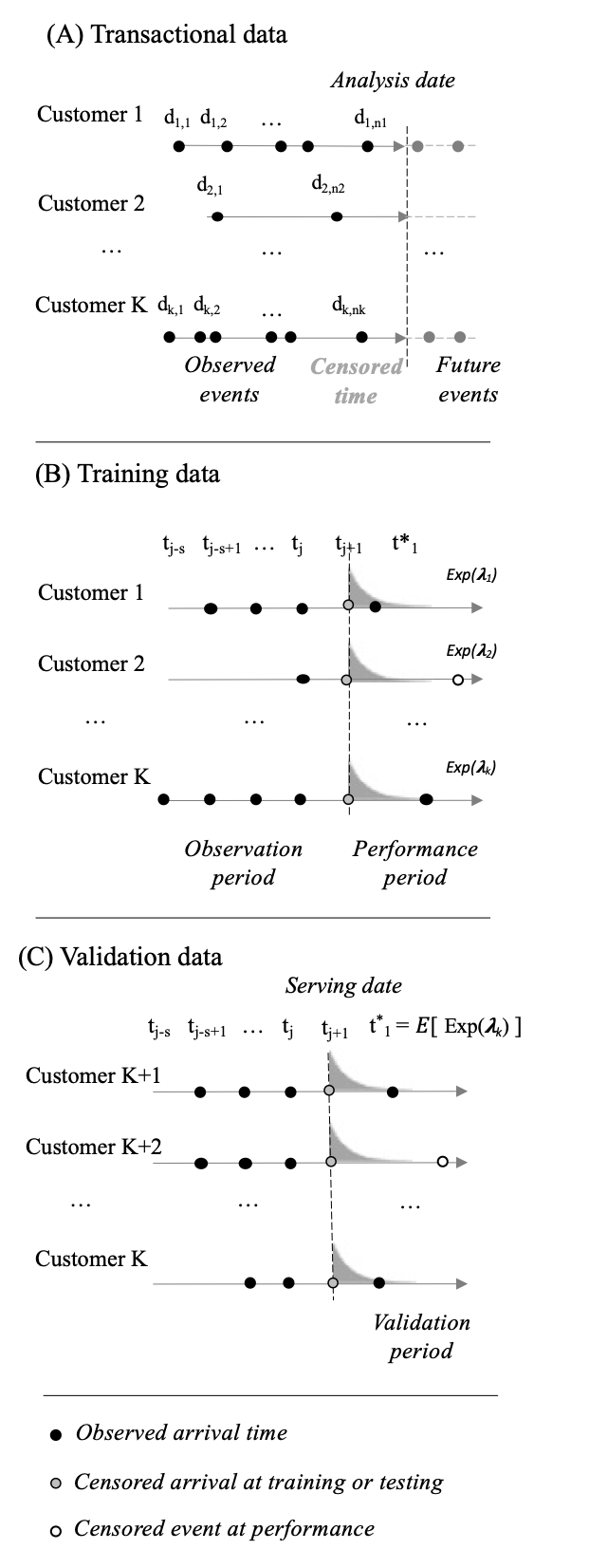}
\caption{Data transformation required to obtain the exponentially distributed $t_{k,j}=d_{k,j}-d_{k,j-1}$. Each customer sequence is observed up to time $j$ to predict $t_{k,j+1}$.}

\label{fig:ITT_Representation}
\end{figure}

 Let’s $D_{K}$ be the set of all purchases made by $K$ customers in a portfolio. In the non-contractual setting, purchases can happen at any date and time, thus, let’s denote the date that customer $k \in K$ made its $i-th$ purchase as $d_{k,i}$ as shown in Figure \ref{fig:ITT_Representation}. Typically, transnational data of all customers is arranged in a single dataset with three columns \emph{'Customer id'}, \emph{'Order Number'}, and \emph{'Purchase date'} as shown in Table  \ref{table:Original_transactional_data}.  This dataset is sorted by each customer id in date-ascending order, which means that $d_{k,i_{1}}\leqslant d_{k,i_{2}}$  for all $i_{1}\leqslant i_{2}$ for a fixed customer $k$. Although some of the time-to-event methods described in section \ref{section2} approach the churn prediction problem with this form of data, in this work it is necessary carry an extra transformation by defining a random variable $T_{k}$ as the time difference between consecutive customer transactions, i.e., $t_{k,i}=d_{k,i}-d_{k,i-1}$ for $i\geqslant 2$. This new random variable $T_{k}$ is assumed to be exponentially distributed at customer level, i.e., $T_{k}\sim Exp(\lambda_{k})$ as illustrated in Figure \ref{fig:itt_distribution}. 

To account for right censoring of purchasing events, we consider the time elapsed since customers made their last purchase against the observation date, and define this distance as $t_{k,n_{k}+1} = analysis \ date - d_{k,n_{k}}$ as the censored time for all customers.

\begin{table}[!htb]
      \caption{Original Transactional data}
      \centering
        \begin{tabular}{ccc}
        \toprule
        Customer ID & Order No. & Purchase date \\ \midrule
        {1} & {0} & $d_{1,0}$ \\ 
        {1} & {1} & $d_{1,1}$ \\ 
        {2} & {0} & $d_{2,0}$ \\ 
        ... & ... & ... \\
        k & i & $d_{k,i}$ \\
        k & i+1 & $d_{k,i+
        1}$ \\
        ... & ... & ... \\ \bottomrule
        \end{tabular}
        \label{table:Original_transactional_data}
\end{table}
\begin{table}[!h]
      \centering
        \caption{Customer sequential-transactions}
            \begin{tabular}{ccc}
            \toprule
            Customer & Arrival-times Sequence  & $\delta_{k}$-sequence\\ \midrule
            1 & [$t_{1,1}, t_{1,2}, ..., t_{1,n_{1}}, t_{1, n_{1}+1}$] & [1,1, ..., 1, 0]\\ 
            2 & [$t_{2,1}, t_{2,2}, ..., t_{2,n_{2}}, t_{2, n_{2}+1}$] & [1,1, ..., 1, 0]\\ 
            %k & [$t_{1,1}, t_{1,2}, ..., t_{1,n_{k}}$] \\ 
            ... & ... & ... \\ 
            k & [$t_{k,1}, t_{k,2}, ..., t_{k,n_{k}}, t_{k, n_{k}+1}$]& [1,1, ..., 1, 0] \\ \bottomrule
            \end{tabular}
            \label{table:Sequential_transactions}
\end{table}

In order to model the $k-th$ customer inter-arrival time $t_{k}$ in a sequential framework, we compress all the arrival times $t_{k,j}$ of each customer into a new sequential vector $\textbf{t}_{k} = [t_{k,1}, t_{k,2}, … , t_{k,n_{k}}]$ for each customer $k \in K$, where each arrival time $t_{k,j}$ is assumed to be exponentially distributed. As expected, each sequence has a different length depending how many purchases customers have made in the company over their lifetime, thus  $length(\textbf{t}_{k}) = n_{k}$, where $n_{k}$ is the total number of purchases made by customer $k$. Then, to consider right censored event times, which are event times that are only partially observed due the fact that customers may make their next purchase after the time of the analysis, we concatenate the time since each customer was last observed with respect to the time of the analysis $t_{k,n_{k}+1}$ at the end of the sequence, and create a binary identifier $\delta_{k,j}$ for each event time in the sequence, which take the values $\delta_{k, j} = 1$ when the event is fully observed and $\delta_{k,j} = 0$ for censored or partially observed times, by construction, the sequence  $\delta_{k}$ for each customer $k$ will only contain one single censored event at the last position of the sequence. Then, the training data $T_{X}$ is defined as the union set of all independent $\textbf{t}_{k}$ vectors, i.e., $T_{X} =  \bigcup_{k \in K} ( \textbf{t}_{k}\cup{t_{k,n_{k}+1})}$  as shown in Table \ref{table:Sequential_transactions}, with their corresponding $\delta_{k,j}$ identifiers. In practice, these vectors $\textbf{t}_{k}$ can be padded at both, training and serving to an arbitrary length vector.

%Survival charts dfg

\subsection*{Training process}
\begin{figure*}[!htbp]
\centering{}
%\captionsetup{width=.8\linewidth}
\includegraphics[scale=0.65]{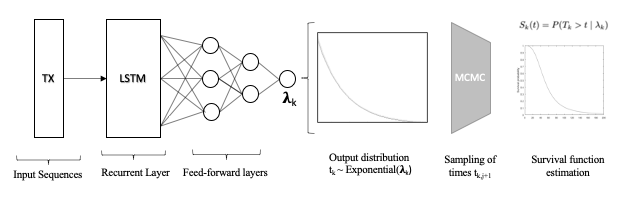}
\caption{Proposed Neural Survival model. The input transactional sequences are passed through a Long-Short Term Memory cell (LSTM) and a Multilayer Perceptron followed by an exponential activation of size 1 to parameterise the customer lever exponential model $\textbf{t}_{k}$. The survival model $S_{k}$ at customer level is drawn from the posterior distribution with Hamilton sampling. }

\label{fig:RNN_model}
\end{figure*}

The training goal is developing a neural network based in machine learning assuming that the arrival times $t_{k}$ are exponentially distributed with some specific parameter $\lambda_{k}$ for each customer $k \in K$. To achieve this, we implement a Recurrent Neural Network followed by a Multilayer Perceptron with a single output unit with sigmoid activation to estimate $\hat{\lambda}_{k} = NN_{T_{X}}(\textbf{t}_{k})$, which is then used to parameterise the exponential density function $g$ of the arrival-times for each customer. At each $j\text{-th}$ time-step the input data for the model is the sequence $\textbf{t}_{k,j} = [t_{k,j-s}, t_{k,j-s+1}, … , t_{k,j}]$ with target $t_{k,j+1}$, where $s$ is the sequence-padding parameter which can be set arbitrary for each application, and it is usually lower than $max(n_{k})$ and fine-tuned during model training. Figure \ref{fig:RNN_model} shows the final model architecture proposed, where estimation of model parameters can be performed in two different ways, firstly, by minimising a weighted asymmetric loss function which considers censored events by using the identifier $\delta_{k}$ created for each observation, with $\delta_{k} = 1$ for fully observed event-times and $\delta_{k} = 0$ for censored times, at the same time of penalising for large predicted values of $\hat{t}$ for censored observations:

% \textcolor{red}{ADD subindex j in the loss}
% \[
% Loss = \left( \frac{1}{N_{\delta=1}}\sum_{{k=1}}^{N}\omega_{t_{k}}(\hat{t_{k}} - t_{k})^2 \cdot 1_{\{\delta_{k}=1\}} \right.
% \]
% \[
% \left. +\frac{1}{N_{\delta=0}}\sum_{{k=1}}^{N}(1-\omega_{t_{k}})(\hat{t_{k}}-E[t])^2 \cdot 1_{\{\delta_{k}=0\}} \right)
% \]
% \medskip

\[
Loss = \left( \frac{1}{K_{\delta=1}}\sum_{{k=1}}^{K}\sum_{{j=1}}^{n_{k}} \omega_{t_{k}}(\hat{t_{k,j}} - t_{k,j})^2 \cdot 1_{\{\delta_{k,j}=1\}} \right.
\]
\[
\left. +\frac{1}{K_{\delta=0}}\sum_{{k=1}}^{K}\sum_{{j=1}}^{n_{k}}(1-\omega_{t_{k}})(\hat{t_{k,j}}-E[t])^2 \cdot 1_{\{\delta_{k,j}=0\}} \right)
\]
\medskip

%$\hat{\lambda} =  \frac{\sum_{k}{\delta_{k}=0}}{\sum_{k}{\delta_{k}=1} \cdot \sum_{k}{t_{k}}} $ 

where the weighting factor $\omega_{t_{k}} = P(\delta_{k} = 1 \mid T_{k}=t)$ represents the probability of fully observed events at time $t_{k}$ in the training data for customer $k$, and $E[t]$ is the estimated expected value of the target distribution of $T$ which can be obtained by estimating the parameter of the exponential distribution under the presence of censored events via maximum log-likelihood in each training batch as 
%  which for simplicity it is assumed to be equal for all observations in each batch during training
\[\hat{\lambda} =  \frac{K_{\delta=1}}{K \cdot \overline{t}}, \]
where $\overline{t}$ is the mean of observed and censored event times \cite{Kalbfleisch_2002}.

Alternatively to the aforementioned asymmetric loss function, as it is commonly seen in similar methods mentioned in the second section, the model can be also trained by maximising the partial likelihood function of the model for censored data given by 
% \[
% L = \prod_{k=1}^K f(t_{k} \mid \hat{\lambda}_{k})^{\delta_{k}} \cdot S(t_{k} \mid \hat{\lambda}_{k})^{1-\delta_{k}} 
% \]

\[
L = \prod_{k=1}^K \prod_{j=1}^{n_{k}} f(t_{k,j} \mid \hat{\lambda}_{k})^{\delta_{k,j}} \cdot S(t_{k,j} \mid \hat{\lambda}_{k})^{1-\delta_{k,j}} 
\]

which is commonly transformed for training into a minimisation task where the objective is minimising the negative log-likelihood for censored events denoted as

% \[
% LL = -\sum_{k}\left(log(f(t_{k}\mid\textbf{t}_{k}))1_{\{\delta_{k}=1\}}+log(S(t_{k} \mid \textbf{t}_{k}))1_{\{\delta_{k}=0\}}\right)
% \]
% \medskip

\begin{equation*}
\resizebox{0.52\textwidth}{!}{$LL = -\sum_{{k=1}}^{K}\sum_{{j=1}}^{n_{k}}\left(log(f(t_{k,j} \mid \hat{\lambda}_{k}))1_{\{\delta_{k,j}=1\}}+log(S(t_{k,j} \mid \hat{\lambda}_{k}))1_{\{\delta_{k,j}=0\}}\right)$}
\end{equation*}

Although the model can be trained with two different methods, by using an asymmetric loss function or maximum likelihood as loss function. For our experiments presented in the fourth section, we decided to train the model by using the negative log-likelihood as loss function, which takes into account that the distribution of errors is not symmetric.  

Finally, at serving phase, the survival probability at time $t$ for a customer $k$ can be easily estimated once $\hat{\lambda}_{k}$ is known, as

% \[
% S_{k}(t) = 1 - P(T_{k}<t) = 1 - \int_{0}^{t}{\lambda_{k}exp(-\lambda_{k}x) dx} = exp(-\lambda_{k}t)
% \]
\begin{equation*}
\begin{aligned}
S_{k}(t) {}& = 1 - P(T_{k}<t) \\ 
& = 1 - \int_{0}^{t}{\lambda_{k}exp(-\lambda_{k}x) dx} \\
& = exp(-\lambda_{k}t)
\end{aligned}
\end{equation*}
\medskip

Then, we define the estimator of $S_{k}(t)$ at customer level by 
\[
\hat{S}_{k}(t) = exp(-\hat{\lambda}_{k}t)
\]

\[
S_{k}(t) = P(T_{k}>t \mid \lambda_{k})
\]

\medskip
Furthermore, to compute the survival status of each customer at an specific serving date we evaluate $\hat{S}_{k}(t)$ at the current recency time of each customer, i.e, the time elapsed between last customer purchase and the serving date. Additionally, it is also possible to obtain a future deferred event probability over a period simply by computing $\hat{S}_{k}(t_{1}) – \hat{S}_{k}(t_{2})$, which is the probability that the next customer purchase will happen between the time interval $(t_{1},  t_{2})$. A pseudo-code description of training and prediction steps is sketched in algorithm \ref{alg:Training_pseudo_code}.

%%% PSEUDO-CODE:
\begin{algorithm}
\caption{Recurrent model pseudo-code}\label{alg:Training_pseudo_code}
\begin{algorithmic}
\State \textbf{Input:} Customer transactional data presented in table 1 including \emph{'Customer id'} and \emph{'Purchase date'}. \\
\State \textbf{Output:} Recurrent Neural Network to predict next customer event time. \\

\State \textbf{Split data into training, testing, and validation datasets:}
\State - Split customer set into two disjoint training and validation sets containing 80\% and 20\% of customer's data respectively.
\State - Split training customer transactions into two disjoint training and testing datasets w.r.t. the selected analysis date.\\

\State \textbf{Data processing:}
\State 1. Obtain time between subsequent purchases for each customer as $t_{k,i}=d_{k,i}-d_{k,i-1}$.
\State 2. Obtain sequence input vectors for each customer as $T_{X} =  \bigcup_{k \in K} \textbf{t}_{k}\cup{t_{k,n'_{k}}}$
\State 3. Pad sequences the same length with a sliding window over each customer sequence and assign the target event time as the immediate next arrival time in the sequence.  \\

\State \textbf{Model Training:}
\State \textbf{Repeat until convergence:}
    \State 1. Initialise recurrent neural network with randomised weights and biases.
    \State 2. Apply network forward pass to the input sequences to obtain the estimated next arrival time or each customer.
    \State 3. Compute the model loss by using the real observed and censored arrival times and the estimated next arrival times.
    \State 4. Update weights and biases in the network.
\\
\State \textbf{Model Prediction:}
\State 1. Create input sequences for validation dataset as $T_{X} =  \bigcup_{k \in K} \textbf{t}_{k}\cup{t_{k,n'_{k}}}$
\State 2. Apply forward pass of the recurrent model to obtain the next estimated time events.
\State 3. Sample estimated next event time $t_{k, j+1}$ from the posterior distribution of event times for each customer to estimate $\hat{\lambda}_{k}$ for each customer
\State 4. Make predictions of estimated customer survival probability by obtaining $\hat{S}_{k}(t) = exp(-\hat{\lambda}_{k}t)$
\end{algorithmic}

\end{algorithm}

\subsection*{Performance Metrics} \label{section3.3}

This section introduces briefly the main methodologies used to evaluate models where censoring is present in the evaluation data. As mentioned previously, accounting for censoring in predicting customer churn induces further challenges in model evaluation overall, as we are not able to fully distinguish customers who are already churned and will not make any further purchase against the ones that are taking a pause between transactions but will return eventually. 

\subsubsection*{Brier Score}

The Brier score introduced by \cite{Graf1999} is a common evaluation metric used in survival analysis to evaluate the accuracy of survival probabilities. It represents the average squared distance between the observed survival status against the predicted survival probability for all subjects. In the absence of censoring, the expected brier score can be obtained as

\[
BS(t)=\frac{1}{K}\sum_{k=1}^K (1_{\{t^*_{k}>t\}}-S_{k}(t))^2,
\]
\medskip
where $t^*_{k}$ represents the first arrival time for customer $k$ in the validation period. However, to account the presence of censoring in survival models, it is necessary to adjust the score with the inverse probability of censoring weights. For each individual it is considered $t'_{k}= min(t_{k}, C_{k})$ along with $\delta^*_{k}=1_{\{t^*_{k} \leqslant C_{k}\}}$, where $C_{k}$ represents the current time under observation for each individual $k$. Let $G(t) = P(C > t)$ be probability of censoring for a time $t$, usually obtained via Kaplan-Meier estimation \cite{Graf1999}. The estimated time-dependant brier score for censored data under the assumption that the event of interest will happen con probability of one if individuals are observed for long enough time is defined as

%\[
%BS(t)=\frac{1}{N}\sum_{k=1}^N \left ( \frac{(0-S_{k}(t))^2\cdot 1_{T\leq t, \delta_{k}=1}}{G(T)}+\frac{(1-S_{k}(t))^2\cdot 1_{T>t}}{G(t)} \right )
%\]

\[
BS(t)=\frac{1}{N}\sum_{k=1}^N \left( \frac{(0-S_{k}(t))^2\cdot 1_{\{t^*_{k}\leq t, \delta^*_{k}=1\}}}{G(t_{k})}\right.
\]
\[\left. +\frac{(1-S_{k}(t))^2\cdot 1_{\{t^*_{k}>t\}}}{G(t)} \right)
\]

%where $\delta_{k}$ is a binary variable that takes the value of 1 if the event of interest occurs within the evaluation period for customer $k$ and 0 otherwise. 
\medskip

% \[
% BS(t)=\frac{1}{N}\sum_{k=1}^N \left( \frac{(0-S_{k}(t))^2\cdot 1_{\{t_{k}\leq t, \delta_{k}=1\}}}{G(t_{k})}\right.
% \]
% \[\left. +\frac{(P(\delta=1 \mid t_{k}>t)-S_{k}(t))^2\cdot 1_{\{t_{k}>t\}}}{G(t)} \right)
% \]

Finally, the Integrated Brier Score (IBS) provides an overall estimation of model performance for all times up to a given time $t_{max}$. 
\[
IBS(t_{max})=\frac{1}{t_{max}}\int_{0}^{t_{max}} BS(t) dt
\]

\subsubsection*{Concordance index}
Harrell’s Concordance index, also known as C-index or C-statistic \cite{Harrell1982}, is one of most used performance metrics for survival models due to its inherent design to account censoring in data. Contrary to metrics that assess the predictive power of a model by measuring the error in predictions, such as brier score, the C-index assess the discriminating power of a risk score by comparing the correlation between predicted scores $\hat{S}(t)$ and true observed times for pairs of comparable individuals $k_{i}$ and $k_{j}$, with $i \ne j$,  who experienced the event at different times $t_{k_{i}}$ and $t_{k_{j}}$ respectively. The C-index is defined as:
%\[
%C-Index = P(S_{k_{1}}(T_{1}) > S_{k_{2}}(T_{2}) \mid T_{1} > T_{2})
%\]

\[
\text{C-Index} = P(S_{k_{j}}(t_{k_{j}}) > S_{k_{i}}(t_{k_{i}}) \mid t_{k_{i}} > t_{k_{j}})
\]

where large values of C-index indicate a more informative prediction of which individuals are more susceptible to experience the event of interest. 

\subsubsection*{Time-dependant Area under the ROC curve}

The receiver operating characteristic (ROC) is a well-established technique in machine learning to assess classification power of binary classifiers, particularly when the time horizon of the target variable is fixed \cite{Fawcett_2006}. For an arbitrary classifier $\hat{Y}:X \rightarrow [0,1]$ with binary outcome $Y = \{0,1\}$ and classification threshold $c$, the ROC curve compares the classifier sensitivity,  obtained as $sensitivity(c) = P(\hat Y_{1}>c \mid Y=1)$ a.k.a True Positive Rate (TPR), against the one minus the specificity, where $specificity(c) = P(\hat Y_{0}\le c \mid Y=0)$ a.k.a. False Positive Rate (FPR), over all possible values of $\hat{Y}$, where $\hat Y_{1}$ is the predicted probability for a positive instance, and $\hat Y_{0}$ is the predicted probability for a negative instance for the defined classification threshold $c$. Thus, the AUC (Area Under the ROC Curve) can be defined as the total area under the ROC curve, and can be interpreted as the probability that a randomly selected pair of observations are correctly classified by $\hat{Y}$ \cite{Kamarudin2017}, i.e :

\[
AUC = P(\hat Y_{1}>\hat Y_{0})
\]
\medskip

AUC is an aggregated performance metric over all possible classification thresholds in a model, a large AUC indicates that the classifier possesses a high predictive power to distinguish between different classes. 

Different methods have been proposed to extend this metric to consider variable time-horizons and probability of event occurrence non-constant, such as in survival models presented in \cite{Kamarudin2017, Heagerty_2005, Lambert_2016}. In simple terms, when extending the ROC curve to a time-dependant outcome, both sensitivity and specificity becomes time-dependant measures with respect to a time-dependant random variable $T$, which are defined by \emph{‘cumulative cases at t’}, individuals who experienced the event before $t$, i.e.,  $t^*_{k} \le t$, and \emph{‘dynamic controls  at t’}, individuals who experienced the event after time $t$, i.e., $t^*_{k}>t$. As such, sensitivity and specificity can be expressed as function of time t \cite{Heagerty_2005} as follows:

\[
sensitivity(c,t) = P(S_{k}(t)>c \mid t^*_{k} \le t)
\]
\[
specificity(c,t) = P(S_{k}(t) \le c \mid t^*_{k}>t)
\]

\medskip

Thus, the Cumulative/Dynamic ROC (C/D ROC) measures how well a model can classify subjects who experienced the event at different points in time, and the Cumulative/Dynamic AUC (C/D AUC) \cite{Heagerty_2000, Hung2010} provides a single aggregated measure of the total area under of the C/D ROC curve, which represents the probability that the estimation of the non-event will be larger for individuals who have already experienced the event at time $t$ compared against those who have not. The estimated C/D AUC is defined at time $t$ as 

\[
AUC(t) = P(S_{k_{i}}(t_{i}) > S_{k_{j}}(t_{j}) \mid t_{i} \le t < t_{j} ) 
%AUC(t) = P(\hat Y_{i}>\hat Y_{j} \mid T_{i} \le t , T_{j} > t), i \ne j
\]

As mentioned previously, most survival analysis techniques assume that the event of interest will happen eventually for all individuals, leading into a potential evaluation bias when considering $AUC(t_{1}, t_{2})$ as performance metric when $t_{2}< \infty$, which happens potentially in every real-world application. Thus, if the event of interest might or might-not happen for all individuals, metrics like Brier score and C-index might be more useful than the time-dependant AUC in survival applications. 

\section*{Experiments and Results} \label{section4}
Profusion is a data consultancy company based in London, UK, that provides data and analytic services to a wide range of businesses in the retail and financial sectors across the UK. Profusion uses different statistical and machine learning methods to identify customers likely to churn in client’s databases. Typically, churn estimation is carried once or twice per month for each company, and as result of this process,  customers likely to churn are targeted in marketing campaigns to proactively encourage them to engage with the company's products and services. 

This work aims to improve classification power at making customer churn prediction for a large retail company in the UK. Due to confidentiality constrains, name of this company will not be shown, and it will be denoted as $company$. Additionally, model performance is also assessed in a synthetic dataset which resembles the main characteristics of transactional data in the retail industry. To compare our methodology, we established two baseline models, an initial transactional only CPH model, and an individual-level Kaplan-Meier estimator.  Although CPH allows to include customer level characteristics as model covariates, such as age, gender, and type of customer, with the purpose of having a fair comparison in model performance at assessing transactional-only input data, the CPH baseline used in this work is based solely in the time elapsed between consecutive events. Similarly to the process presented previously, a censored event time obtained from the distance between customer last purchase and analysis dates is assumed for each customer. 

As mentioned, CPH does not model directly the event time, but the hazard function of individuals at time $t$, which is the probability of individuals experiencing the event of interest at time $t$, and once the hazard function is known the survival function of individuals can be estimated as $\hat{S}(t)=\text{exp}(-\Gamma(t))$, where $\Gamma(t)$ is the cumulative hazard function.

Additionally, we compare model performance against an individual Kaplan-Meier baseline, for this, we estimated the survival function $\hat{S}_{k}$ for each customer by considering all the observed and censored event times for customer $k$, and made predictions of its current survival status with respect to its current censoring time.

\subsection*{Dataset 1: \emph{Retail data}}
\begin{figure}[!h]
\centering{}
\includegraphics[scale=0.45]{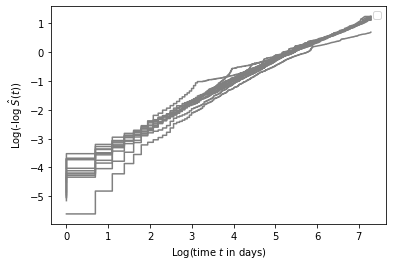}
\caption{Estimated survival function of retail dataset via Kaplan-Meier curve for 12 monthly cohorts of customers who made a purchase in the last 30 days.}
\label{fig:Monthly_survival_distributions_ds1}
\end{figure}

The first dataset contains transactional data of $company$, a large retailer in the UK with almost 200 physical stores and online ordering. Between 29/03/2016 and 20/04/2021, the $company$ had a total of 17.8 million transactions made by 2.8 million customers, with an average time between observed events of 80 days for customers with at least 2 purchases, i.e., the estimated $\lambda$ parameter for the time-to-event exponential model considering all customers is $\hat{\lambda} = 0.012$. Although some demographic information about individual customers is available, such as type of customer and location, in order to model customer churn as described in the third section, we only include the customer id and the purchase date in our analysis to obtain times between events. Due to the non-contractual setting of the retail industry, this dataset presents an estimated monthly customer drop-out of 14.7\% after 1400 days, i.e., 14.7\% of customers will not make any further purchase in the company. Figure \ref{fig:Monthly_survival_distributions_ds1} shows the cumulative logarithmic survival probability of this dataset for a 12 different months sliding-window. 

Finally, in model training we consider all transactions made by a random sample of 100,000 customers.

\subsection*{Dataset 2: \emph{Synthetic data}}

\begin{figure}[!h]
\centering{}
\includegraphics[scale=0.45]{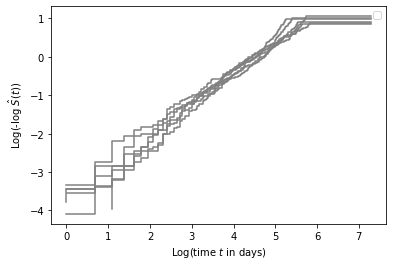}
\caption{Estimated survival function of synthetic dataset via Kaplan-Meier curve for 12 monthly cohorts of customers who made a purchase in the last 30 days.}
\label{fig:monthly_survival_curves_ds2}
\end{figure}

Following \cite{Bender_2005} a synthetic dataset of realistic multi-event survival data with known $\lambda_{k}$ for each individual can be obtained. To obtain a large enough dataset, we simulate a set of 100,000 customers where $\lambda_{k}$ for each customer is drawn from a Gaussian distribution with mean $\mu =  0.08$, and standard deviation $\sigma=0.02$, i.e., the expected customer return time for this synthetic dataset is every 12.5 days. To avoid negative values of $\lambda_{k}$ we specify a minimum threshold of 0.01, thus, $\lambda_{k} = max(0.01, N(\mu, \sigma^2))$. Then, to introduce the effect of customers not making any further purchase due to the inherent non-contractual setting present in the retail industry, a stopping probability of 15\% is introduced for every customer at each sampling iteration of $t_{k,j} \sim exp(\lambda_{k})$. For this dataset in which the estimated monthly customer drop-out is 8.7\% after 120 days. Figure \ref{fig:monthly_survival_curves_ds2} shows the cumulative logarithmic survival probability of this dataset for a 6 different months sliding-window. And Table \ref{table:Main_statistics} presents a list of main summary statistics for both experimental datasets, including dataset size, frequency of the events in the data, training and performance periods, and testing split size. 

\subsection*{Results}
As stated previously, the time between purchases is exponentially distributed for both datasets analysed. In all cases, the probability of a customer making its next purchase decreases significantly as time passes, and just few transactions have happened after a period of 100 days after the last purchase. Table \ref{table:Main_statistics} presents a list of summary statistics for each datasets.

The data pre-processing for all datasets is carried as stated in the third section. For each dataset, we compress the transactional data prior to an arbitrary analysis date into customer level sequential representations to obtain a dataset with a similar form than in table \ref{table:Sequential_transactions}, i.e., for each dataset we create the training data as $T_{X} =  \bigcup_{K} [t_{k, n_{k}-s}, t_{k, n_{k}-s+1}, ..., t_{k, n_{k}}\cup{t_{k,n'_{k}}}]$ for every $k \in K$, where $s$ is the sequence-padding parameter used to consider all sequences to be of the same length at training, these padding parameters are shown in table \ref{table:Main_statistics} for each dataset. As it is commonly done in modelling sequential data, the analysis date for training should be chosen in a way that the performance period excludes completely the first date available for serving. In our experiments, analysis dates for all dataset are set to at least 6 months before the model serving date, besides we specify a validation set of customers completely disjoint to the customers used at training for each model. Figure \ref{fig:ITT_Representation} shows a visual representation of the overall data processing needed to create the the vectors $\textbf{t}_{k}$ for each customer. 

\begin{table*}[ht]
\caption{Main summary statistics for both experimental datasets.}
\centering
\begin{tabular}{lcc}
\toprule
 & Retail Dataset & Synthetic Dataset \\ \midrule
Number of customers & 2.8 M  & 100 K \\
Training size & 10\%  & 80\% \\
Validation size & 5\%  & 20\% \\
Observation period (MM/YYYY) & 03/2016 - 05/2020 &  N/A \\
Performance period (MM/YYYY) & 06/2020 - 04/2021 &  N/A \\
Median customer purchases  & 2 & 12\\
Median time between purchases & 32 days & 12.5 days \\
Length of padded sequences (s) & 5 & 7 \\\bottomrule
\end{tabular}
\label{table:Main_statistics}
\end{table*}

\begin{figure*}[!htbp]
\centering{}
\includegraphics[scale=0.55]{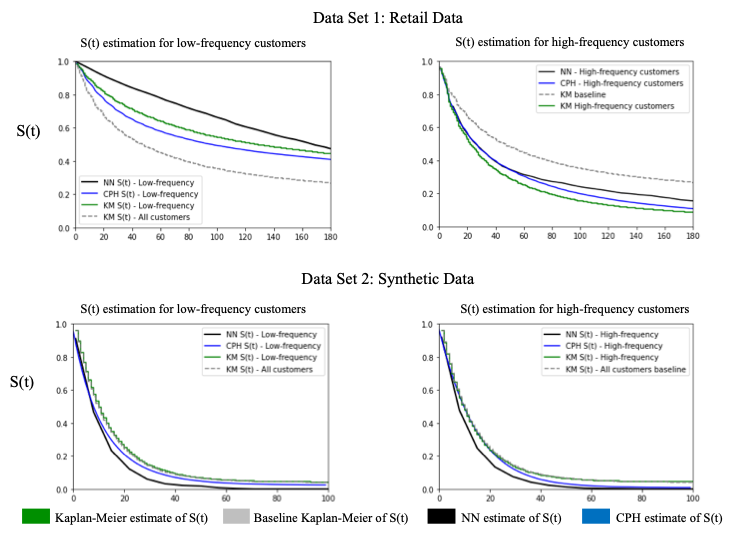}
\caption{Estimated function $S(t)$ for an average customer in  both validation datasets with respect to low and high customer purchase frequency.}
\label{fig:survival_all_data}
\end{figure*}

\begin{figure*}[!htbp]
\centering{}
\includegraphics[scale=0.55]{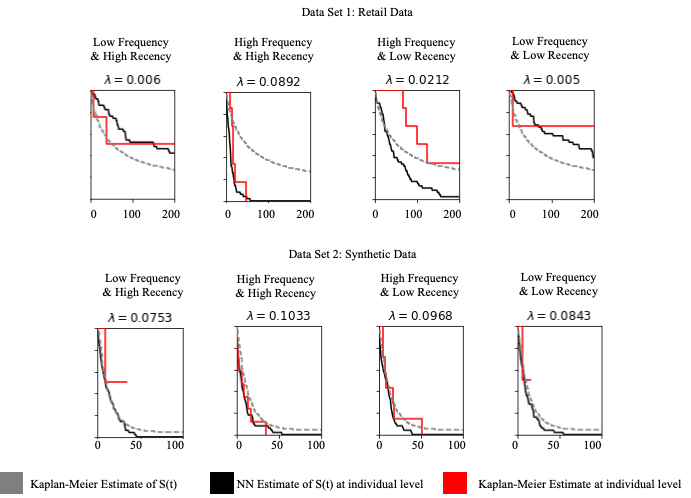}
\caption{Estimated survival function at customer level up to 200 and 100 days for customers belonging to different groups in the dataset with respect to their frequency of purchase and current recency at the time of the analysis. }

\label{fig:Survival_curves_15_customers}
\end{figure*}

At prediction time, we create the input sequences $[t_{k, n_{k}-s}, t_{k, n_{k}-s+1}, ..., t_{k, n_{k}}, t_{k, n_{k}+1}]$ as described in the third section to estimate $\hat{\lambda}_{k}$ for each customer $k \in K$ in the performance dataset, which is set to be the complement of the training dataset for both experiments. Then, the survival probabilities $S_{k}(T>t)$ for any time $t$ can be estimated by integrating the exponential model parameterised by $\hat{\lambda}_{k}$ for each customer. Figure \ref{fig:Survival_curves_15_customers} shows the estimated survival curves obtained for individual customers belonging to four different segments with respect to their frequency of purchase at the time of the analysis. Figure \ref{fig:survival_all_data} shows the overall estimation of the survival function for customers with high and low purchase frequency in the corresponding validation datasets using all three different methods to estimate survival probabilities. Additionally, Table \ref{table:Performance_results} shows the performance results obtained in both datasets of the metrics mentioned previously in the third section, as shown, the recurrent model can outperform both baseline methods in terms of the brier score, however, both the C-Index and the time-dependant AUC are significantly higher for the Cox model.

\begin{figure}[!h]
\centering{}
\includegraphics[scale=0.42]{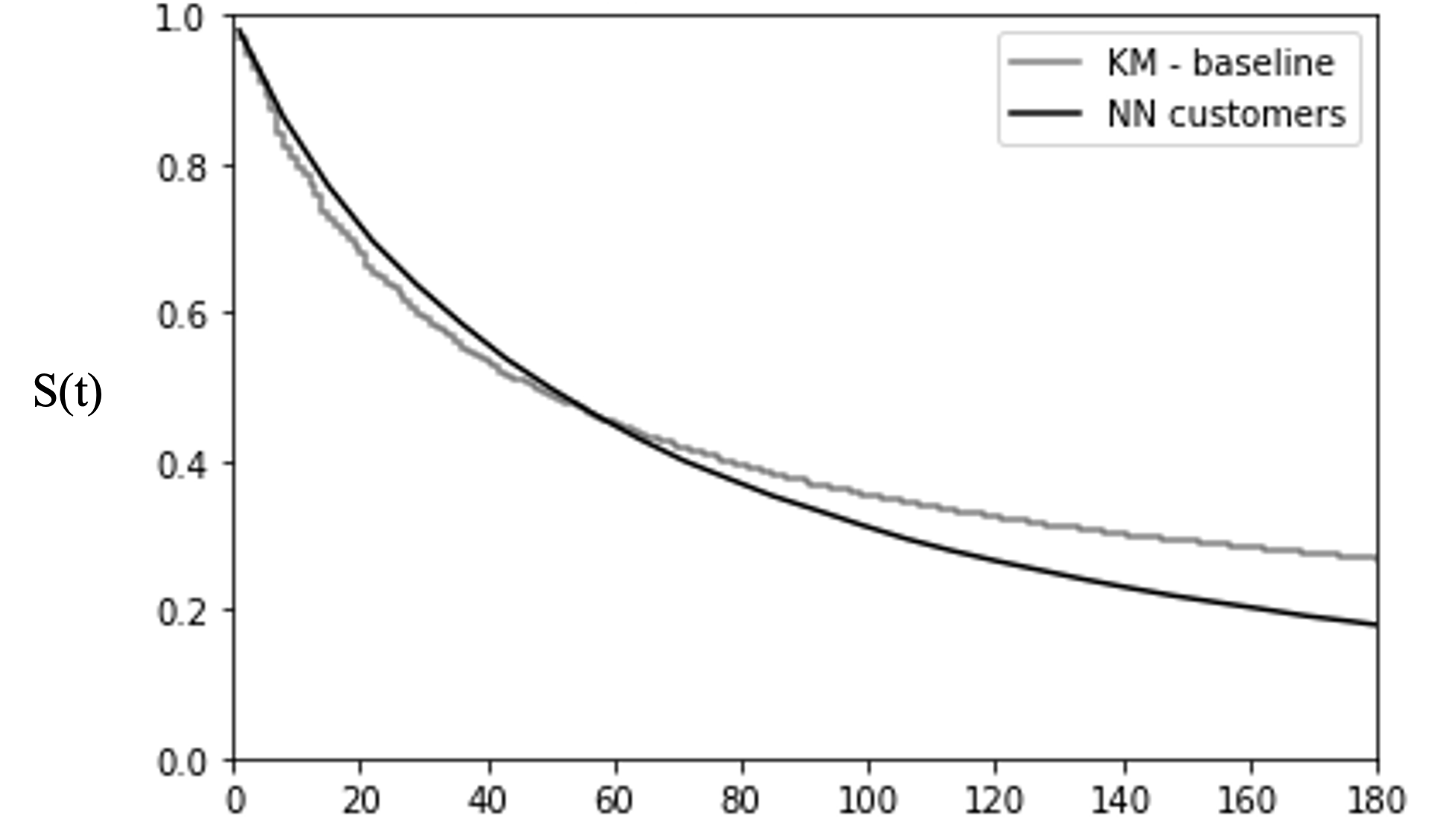}
\caption{Point-wise average of estimated survival functions $S(t)$ at customer level by proposed neural network and baseline Kaplan-Meier estimator.}
\label{fig:KM_and_NN_survival_function}
\end{figure}

As expected, it is seen from Figure \ref{fig:survival_all_data} and Figure \ref{fig:KM_and_NN_survival_function} that the Recurrent Neural Network model can learn somehow efficiently a survival distribution of event times close to the Kaplan-Meier estimate of $S(t)$, although due to the exponentially distributed assumption under the event times distribution, the overall survival function $S(t)$ estimated with the NN will never match perfectly the Kaplan-Meier estimated survival function. As the synthetic dataset was designed in a way that frequency of events and recency (time since last event) do not affect customer's churn rate, it is expected to not see a large impact in the estimated survival curves for customers with different frequency of purchase, as it is shown in Figure \ref{fig:survival_all_data} for synthetic dataset.

\begin{figure}[!h]
\centering{}
\includegraphics[scale=0.42]{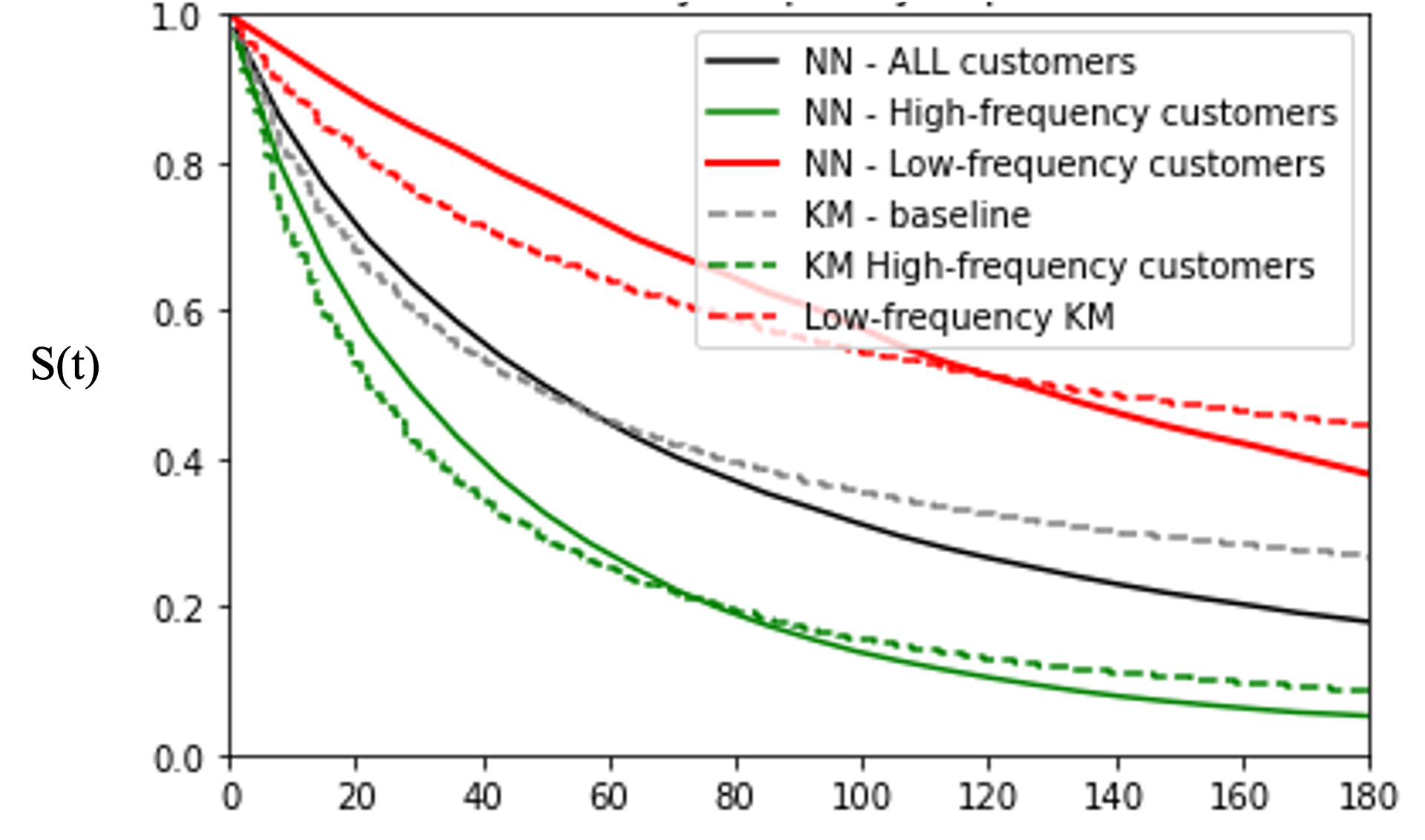}
\caption{Point-wise average of estimated survival functions $S(t)$ at customer level by proposed neural network and baseline Kaplan-Meier estimator with respect to high and low frequency of purchase.}
\label{fig:density_of_lambdas_by_scope}
\end{figure}

\begin{figure}[!h]
\centering{}
\includegraphics[scale=0.45]{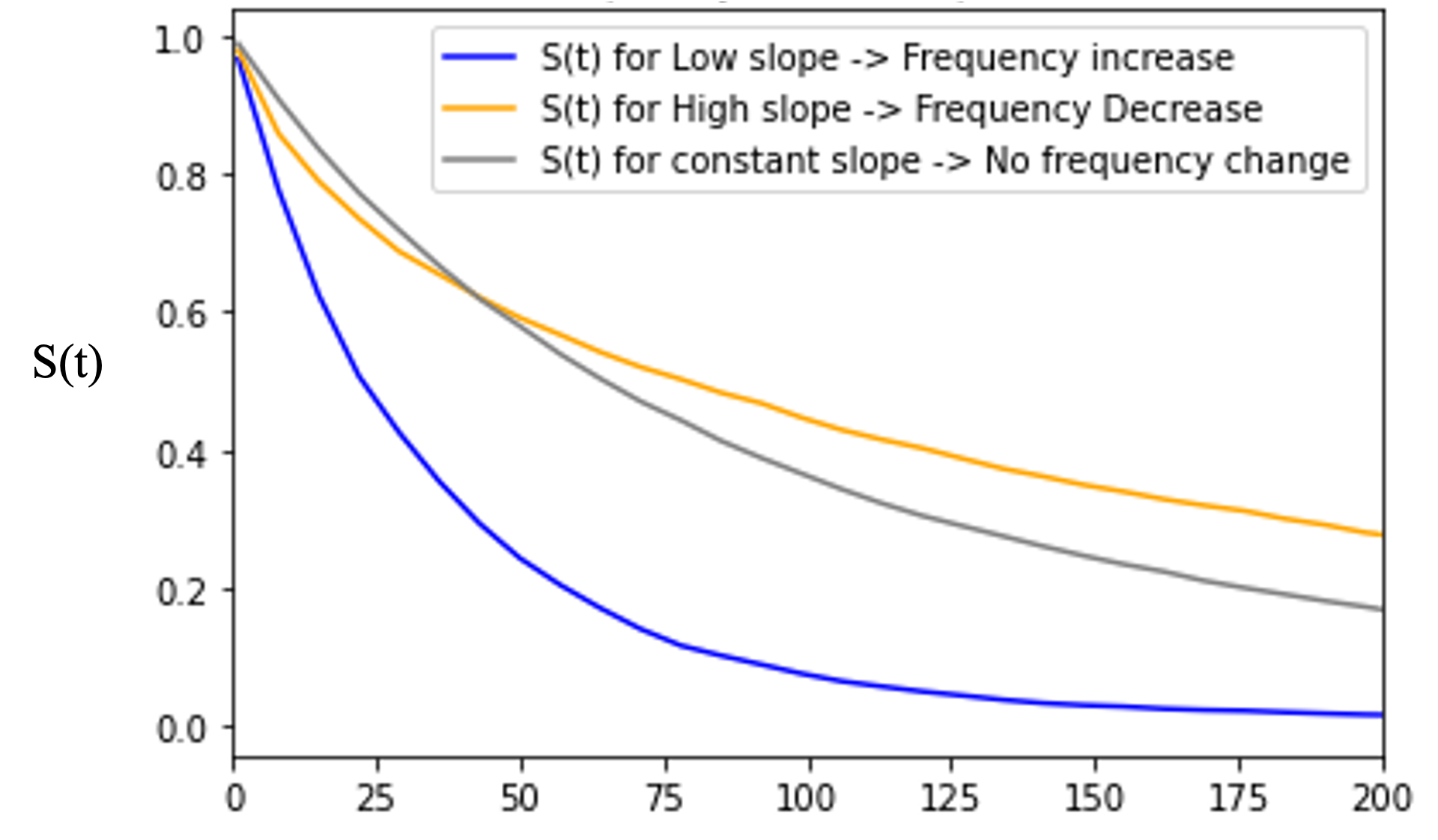}
\caption{Point-wise average of estimated survival function of retail dataset by customers with observed change in frequency of purchase.}
\label{fig:survival_estimation_high_low_slope}
\end{figure}

Additionally, we carried an analysis with the retail dataset to analyse how our method using neural network to estimate survival probabilities captures changes in the frequency of purchases for single customers. For this, we analyse the slope of a linear regression obtained from the event times of each single customer, and check whether this slope is high, indicating that the time to events increase and the frequency of purchase decreases, low, indicating that the time to events decreases and the frequency of purchase increases, and constant, indicating that there is no change in frequency of purchase with the sequence. Figure \ref{fig:density_of_lambdas_by_scope} shows the estimated lambda distribution for these three different groups of customers, and Figure \ref{fig:survival_estimation_high_low_slope} shows the estimated survival distribution $S(t)$ for customers belonging to these three groups of customers. 

\begin{figure}[!h]
\centering{}
\includegraphics[scale=0.45]{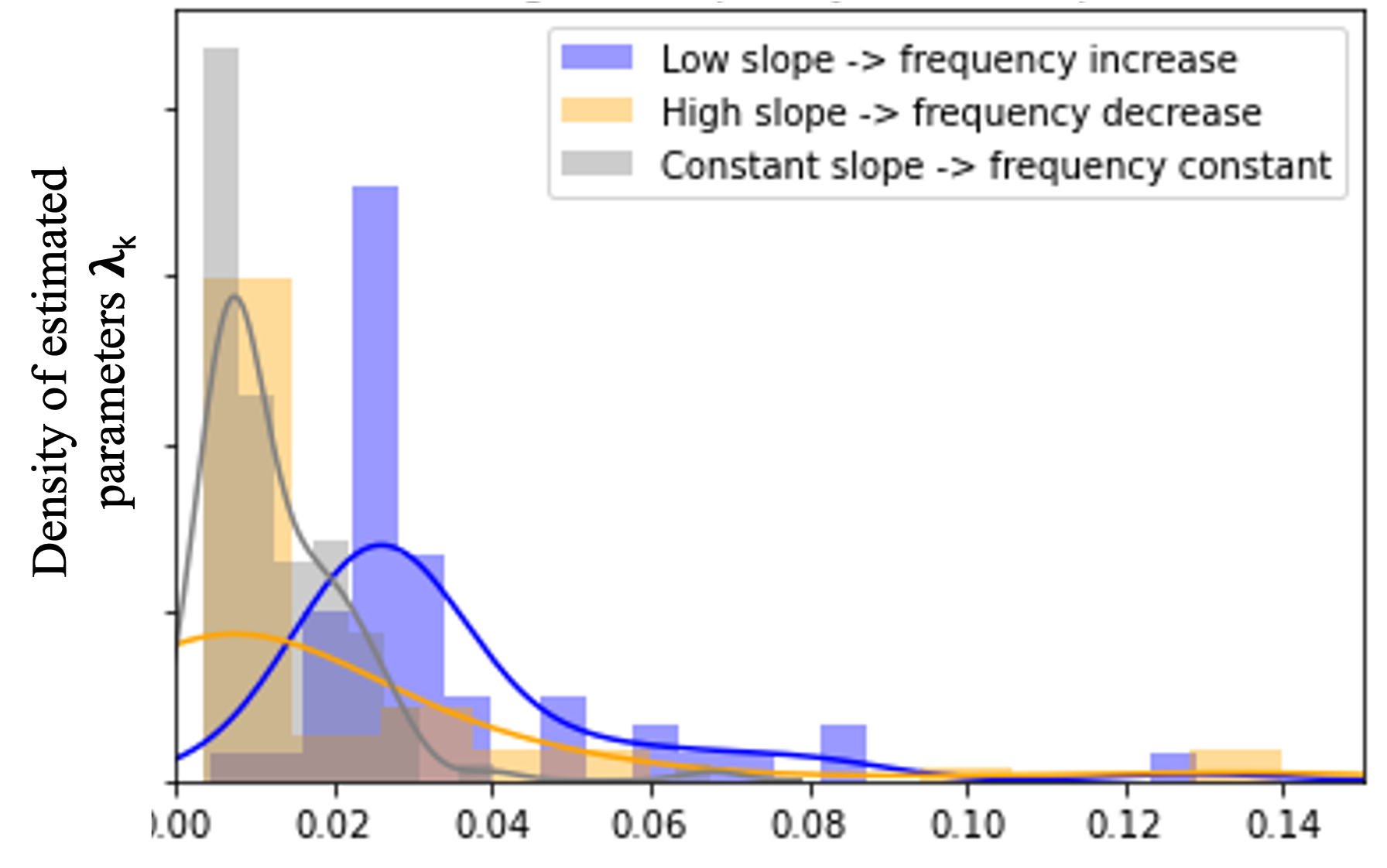}
\caption{Distribution of estimated parameters $\lambda_{k}$ in retail dataset by customers with observed change in frequency of purchase.}
\label{fig:density_of_lambdas_by_scope}
\end{figure}

Finally, we evaluate models’ performance in terms of accuracy, sensitivity and specificity by assessing models as binary classification tasks, where the goal is using the estimated survival probability of customers to predict if the event occurs before time $t$.  For this, we compute the median survival probability of the event at time $t$, denoted by $m(t)$ and use a classification rule that predicts a positive event for observations with $S_{k}(t)< m(t)$, i.e., the event might happen sooner than later for customer $k$, and a negative prediction otherwise.  Table \ref{table:Performance_results_binary_classification} shows results obtained for this evaluation.

\begin{table}[!htbp]
\small
\caption{Model performance obtained in the experiments carried for both datasets in testing and validation splits. Large C-index, large time-dependant AUC, and small Brier score indicate high model performance.}
\resizebox{0.47\textwidth}{!}{
\centering
\begin{tabular}{lcccc}
 & \multicolumn{2}{c}{Retail data} & \multicolumn{2}{c}{Synthetic data} \\ 
 \midrule
 & Test & Validation & Test & Validation \\  
 \midrule
  &  &  &  \\
 & \multicolumn{4}{c}{\textbf{Brier Score at 180 and 100 days}} \\ \midrule
 %Brier score
RNN & 0.303 & 0.369 & 0.350 & 0.431 \\
CPH & \textbf{0.070} & \textbf{0.197} & \textbf{0.021} & \textbf{0.186} \\ 
Ind - KM & 0.384 & 0.437 & 0.378 & 0.292 \\
 &  &  &  \\
 & \multicolumn{4}{c}{\textbf{C-Index}} \\ \midrule
 % C-index
RNN & 0.391 & 0.545 & 0.500 & \textbf{0.498} \\ 
CPH & \textbf{0.580} & 0.233 & \textbf{0.675} & 0.421 \\
Ind - KM & 0.555 & \textbf{0.574} & 0.495 & 0.465 \\
 &  &  &  \\
 & \multicolumn{4}{c}{\textbf{Time-dependant AUC}} \\ \midrule
 % AUC
RNN & 0.348 & 0.458 & 0.487 & \textbf{0.493} \\
CPH & 0.259 & 0.307 & 0.364 & 0.482 \\ 
Ind - KM & \textbf{0.591} & \textbf{0.590} & \textbf{0.494} & 0.468 \\ \bottomrule
\end{tabular}}

\label{table:Performance_results}
\end{table}

\begin{table}[!htbp]
\small
\caption{Model performance obtained in the experiments carried for both datasets considering using the expected survival function to predict whether the event will happen before or after a time $t$, $t=180\text{ days}$ and $t=100\text{ days}$  for retail and synthetic data respectively. As a binary classification problem, larger values of accuracy, sensitivity and specificity indicate higher model prediction capabilities.}
\resizebox{0.47\textwidth}{!}{
\centering
\begin{tabular}{lcccc}
 & \multicolumn{2}{c}{Retail data} & \multicolumn{2}{c}{Synthetic data} \\ 
 \midrule
 & Test & Validation & Test & Validation \\  
 \midrule
  &  &  &  \\
 & \multicolumn{4}{c}{\textbf{Accuracy}} \\ \midrule
% Accuracy
RNN & \textbf{0.654} & 0.525 & 0.560 & 0.210 \\
CPH & 0.485 & 0.462 & 0.366 & \textbf{0.511} \\ 
Ind - KM & 0.592 & \textbf{0.646} & \textbf{0.712} & 0.322 \\
 &  &  &  \\
 & \multicolumn{4}{c}{\textbf{Sensitivity}} \\ \midrule
 % Recall = Sensitivity
RNN & 0.659 & 0.457 & \textbf{0.883} & \textbf{0.833} \\ 
CPH & 0.539 & 0.541 & 0.491 & 0.427 \\
Ind - KM & \textbf{0.664} & \textbf{0.616} & 0.780 & 0.769 \\
 &  &  &  \\
 & \multicolumn{4}{c}{\textbf{Specificity}} \\ \midrule
 % Specificity
RNN & 0.325 & 0.683 & 0.147 & 0.115 \\
CPH & 0.419 & 0.436 & 0.146 & \textbf{0.623} \\ 
Ind - KM & \textbf{0.560} & \textbf{0.718} & \textbf{0.258} & 0.227 \\

\bottomrule
\end{tabular}}
\label{table:Performance_results_binary_classification}
\end{table}

\section*{Conclusion}
Companies around the world are interested in knowing which customers are likely to churn in order to make proactive retention efforts in keeping them engaged with the brand and incentive interaction between customers and products. By predicting the probability of customers making their next purchase over time, our model is capable of estimating the individual-level survival function for each customer instead of an overall survival model for the entire population.

Using recurrent neural networks in time-to-event modelling to predict customer churn allows to model customer purchasing behaviour entirely from transactional data, leaving aside all customer level characteristics, such as age, gender, and income, which are commonly used by companies to estimate how likely is a customer to engage with the brand, and therefore, to purchase again. Additionally, by modelling the time-to-purchase as a sequential problem with a recurrent network architecture, such as the LSTM, the model can learn dependencies in historical interactions to match or improve churn prediction performance of well-established survival techniques with a minimum effort in performing a feature engineering phase or obtaining expensive hand-crafted characteristics from the input data.

However, our approach also has its limitations, firstly, assuming an exponential distribution over the event times for every customer can potentially lead into an underestimation of the survival probability remaining at time $t$ for some the most loyal segments of customers, in which the probability of the next purchase should remain high even after long observation periods without purchases. Secondly, our method requires a large number of purchases made by each customer to provide reliable predictions and the model might be less accurate in customer with short purchasing history, as seen in the experimental results, the model achieved better performance for the synthetic dataset, in which the frequency of purchases is considerably larger than in the retail data. Therefore, this method is more suitable for retail companies where the frequency of customer purchases is high. Nevertheless, our methods is capable to provide estimation of customer churn status and survival probability at individual level for customers with only few event times, which is not possible or not using other methods such as individual Kaplan-Meier, or not accurate in methods such as CPH.

Future work can explore the generalization of this method by combining multivariate time series from different signals as input data for the LSTM layers in the model, as well as compressing seasonal information of purchases or incorporating context information about the purchased items, which is most often easy available in the retail industry.

% \[
% S_{k}(t)=P(T_{k}>t \mid \textbf{t}_{k})
% \]

\backmatter
\newpage
\section*{Declarations}

\bmhead{Funding}
This work was supported by the Knowledge Transfer Partnership program through Innovate UK, University of Essex (Registration Number: Z699129X) and Profusion Media LTD (a company registered in England, number 6947442).

\bmhead{Competing interests}
The authors have no relevant financial or non-financial interests to disclose.

% \bmhead{Ethics approval}
% Not Applicable.

% \bmhead{Consent to participate}
% Not Applicable.

% \bmhead{Consent for publication}
% Not Applicable.

% \bmhead{Availability of data and materials}
% Not Applicable.

\bmhead{Code availability}
The code for this work is intellectual property of Profusion Media LTD. and not publicly available. 

\bmhead{Authors' contributions}
J.E implemented the algorithms, conducted the data analysis and wrote the initial draft, and B.L. verified the results and supervised this work. All authors revised the paper. All authors read and agreed to the published version of the manuscript.

% [CHECK DECLARATION AT Deep Learning-Based Survival Analysis for High-Dimensional Survival Data]

%\begin{itemize}
%\item Conflict of interest/Competing interests (check journal-specific guidelines for which heading to use)
%\item Ethics approval 
%\item Consent for publication
%\item Availability of data and materials
%\item Authors' contributions
%\end{itemize}

%\noindent
%If any of the sections are not relevant to your manuscript, please include the heading and write `Not applicable' for that section. 

\bigskip
\newpage

\bibliographystyle{plain}
\bibliography{biblio}% common bib file


\end{document}